\documentclass[11pt,letterpaper]{article}
\usepackage{emnlp2016}
\usepackage[font=footnotesize,labelfont=bf]{caption}
\usepackage{amssymb}
\usepackage{times}
\usepackage{booktabs}
\usepackage{latexsym}
\usepackage{graphicx}
\usepackage{multirow}
\usepackage{enumitem}
\usepackage{url}
\usepackage{amsmath}

\emnlpfinalcopy

\def\emnlppaperid{***}

\newcommand{\mb}{\mathbf}

\makeatletter
\def\citealt{\def\citename##1{{\frenchspacing##1}, }\@internalcitec}

\def\@citexc[#1]#2{\if@filesw\immediate\write\@auxout{\string\citation{#2}}\fi
  \def\@citea{}\@citealt{\@for\@citeb:=#2\do
    {\@citea\def\@citea{;\penalty\@m\ }\@ifundefined
       {b@\@citeb}{{\bf ?}\@warning
       {Citation `\@citeb' on page \thepage \space undefined}}%
{\csname b@\@citeb\endcsname}}}{#1}}

\def\@internalcitec{\@ifnextchar [{\@tempswatrue\@citexc}{\@tempswafalse\@citexc[]}}

\def\@citealt#1#2{{#1\if@tempswa, #2\fi}}
\makeatother

\title{Cultural Shift or Linguistic Drift? Comparing Two \\Computational Measures of Semantic Change}

\author{William L.\@ Hamilton, Jure Leskovec, Dan Jurafsky \\
Department of Computer Science, Stanford University, Stanford CA, 94305\\
\texttt{wleif,jure,jurafsky@stanford.edu}}

\date{}

\begin{document}
\maketitle

\begin{abstract}
Words shift in meaning for many reasons, including 
cultural factors like new technologies and 
regular linguistic processes like subjectification.
Understanding the evolution of language and culture requires 
disentangling these underlying causes. 
 Here we show how two different distributional measures can be used to detect two different types of semantic change. 
 The first measure, which has been used in many previous works, analyzes global shifts in a word's distributional semantics; it is sensitive to changes due to regular processes of linguistic drift, such as the semantic generalization of \textit{promise} (``I promise.''${\rightarrow}$``It promised to be exciting.'').
 The second measure, which we develop here, focuses on local changes to a word's nearest semantic neighbors; it is more sensitive to cultural shifts, such as the change in the meaning of \textit{cell} (``prison cell'' ${\rightarrow}$ ``cell phone'').
 Comparing measurements made by these two methods allows researchers to determine whether changes are more cultural or linguistic in nature, a distinction that is essential for work in the digital humanities and historical linguistics. 
\end{abstract}

\section{Introduction}
\begin{figure*}
\includegraphics[width=\textwidth]{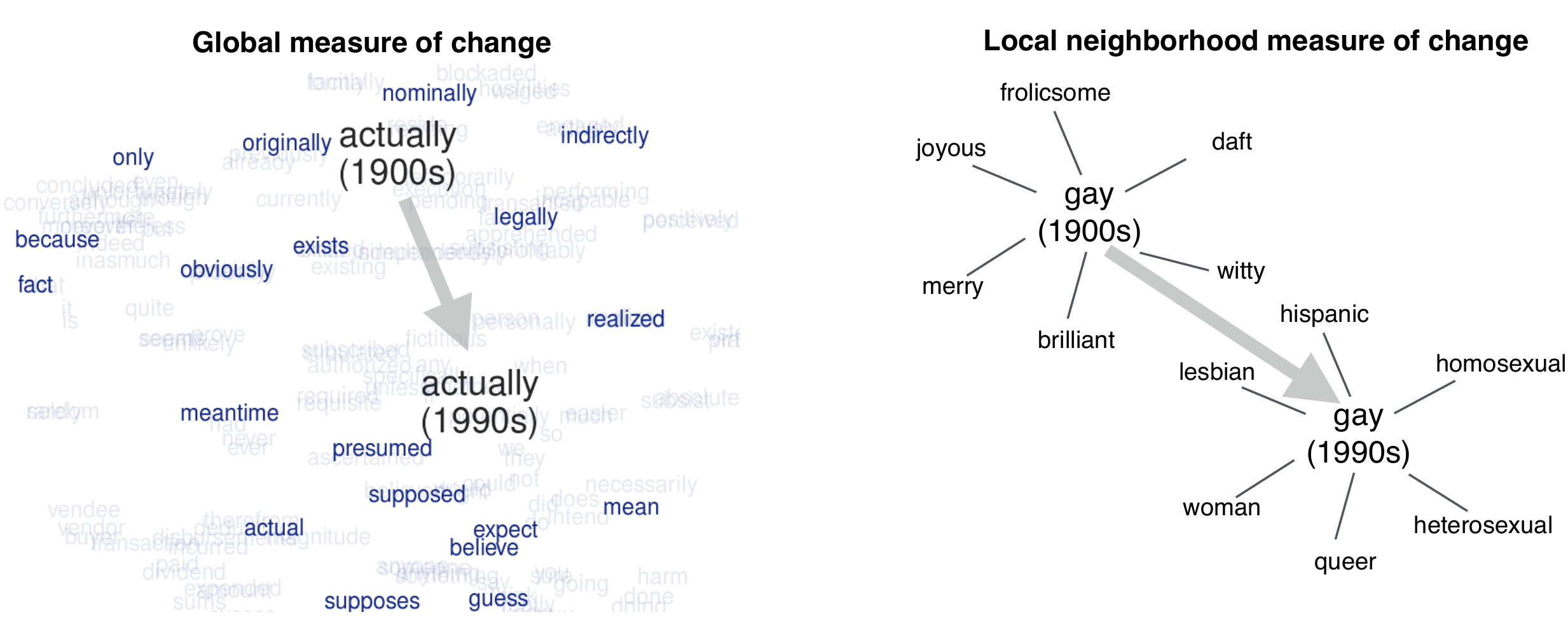}
\caption{{\footnotesize \textbf{Two different measures of semantic change.} With the global measure of change, we measure how far a word has moved in semantic space between two time-periods. This measure is sensitive to subtle shifts in usage and also global effects due to the entire semantic space shifting. For example, this captures how \textit{actually} underwent subjectification during the 20th century, shifting from uses in objective statements about the world (``actually did try'') to subjective statements of attitude (``I actually agree''; see \protect\citealt{traugott_regularity_2001} for details).
In contrast, with the local neighborhood measure of change, we measure changes in a word's nearest neighbors, which captures drastic shifts in core meaning, such as \textit{gay}'s shift in meaning over the 20th century.}
}
\label{measures}
\vspace{-5pt}
\end{figure*}

Distributional methods of embedding words in vector spaces according to their co-occurrence statistics are a promising new tool for diachronic semantics 
\cite{gulordava_distributional_2011,jatowt_framework_2014,kulkarni_statistically_2014,xu_computational_2015,hamilton_diachronic_2016}.
Previous work, however, does not consider the underlying causes of
semantic change or how to distentangle different types of change.

We show how two computational measures can be used to distinguish between semantic changes caused by cultural shifts (e.g., technological advancements) and those caused by more regular processes of semantic change (e.g., grammaticalization or subjectification). 
This distinction is essential for research on linguistic and cultural evolution.
Detecting cultural shifts in language use is crucial to computational studies of history and other digital humanities projects. By contrast, for 
advancing  historical linguistics, cultural shifts amount to noise and only the more regular shifts matter.  

Our work builds on two intuitions:  that
distributional models can highlight syntagmatic versus paradigmatic
relations with neighboring words
\cite{schutze_vector_1993} and 
that nouns are more likely to undergo changes due to irregular cultural shifts while verbs more readily participate in regular processes of semantic change \cite{gentner_verb_1988,traugott_regularity_2001}.
We use this noun vs.\@ verb mapping as a proxy to compare our two measures' sensitivities to cultural vs.\@ linguistic shifts. 
Sensitivity to nominal shifts indicates a propensity to capture irregular cultural shifts in language, such as those due to technological advancements \cite{traugott_regularity_2001}.
Sensitivity to shifts in verbs (and other predicates) indicates a propensity to capture regular processes of linguistic drift \cite{gentner_verb_1988,kintsch_metaphor_2000,traugott_regularity_2001}.

The first measure we analyze is based upon changes to a word's local semantic neighborhood; we show that it is more sensitive to changes in the nominal domain and captures changes due to unpredictable cultural shifts.
Our second measure relies on a more traditional global notion of change; we show that it better captures changes, like those in verbs, that are the result of regular linguistic drift.

Our analysis relies on a large-scale statistical study of six historical corpora in multiple languages, along with case-studies that illustrate the fine-grained differences between the two measures. 

\section{Methods}

We use the diachronic word2vec embeddings constructed in our previous work \cite{hamilton_diachronic_2016} to measure how word meanings change between consecutive decades.\footnote{\scriptsize\url{http://nlp.stanford.edu/projects/histwords/}. This URL also links to detailed dataset descriptions and the code needed to replicate the experiments in this paper.} 
In these representations each word $w_i$ has a vector representation $\mb{w}^{(t)}$ \cite{turney_frequency_2010} at each time point, which captures its co-occurrence statistics for that time period. 
The vectors are constructed using the skip-gram with negative sampling (SGNS) algorithm \cite{mikolov_distributed_2013} and post-processed to align the semantic spaces between years. 
Measuring the distance between word vectors for consecutive decades allows us to compute the rate at which the different words change in meaning \cite{gulordava_distributional_2011}. 

We analyzed the decades from  1800 to 1990 using vectors derived from the Google N-gram datasets \cite{lin_syntactic_2012} that have large amounts of historical text (English, French, German, and English Fiction). 
We also used vectors derived from the Corpus of Historical American English (COHA), which is smaller than Google N-grams but was carefully constructed to be genre balanced and contains word lemmas as well as surface forms \cite{davies_corpus_2010}.
We examined all decades from 1850 through 2000 using the COHA dataset and used the part-of-speech tags provided with the corpora.

\subsection{Measuring semantic change}
\begin{figure*}
\centering
\includegraphics[width=0.9\textwidth]{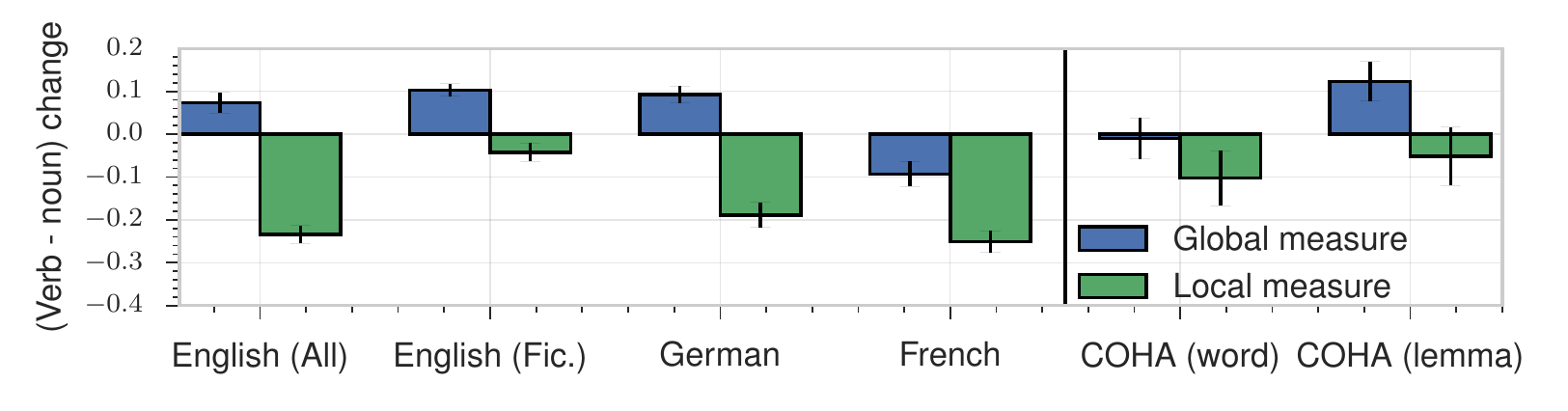}
\vspace{0pt}
\caption{\textbf{The global measure is more sensitive to semantic changes in verbs while the local neighborhood measure is more sensitive to noun changes.} Examining how much nouns change relative to verbs (using coefficients from mixed-model regressions) reveals that the two measures are sensitive to different types of semantic change. Across all languages, the local neighborhood measure always assigns relatively higher rates of change to nouns (i.e., the right/green bars are lower  than the left/blue bars for all pairs), though the results vary by language (e.g., French has high noun change-rates overall). $95\%$ confidence intervals are shown.}
\label{pos-effects}
\vspace{0pt}
\end{figure*}

We examine two different ways to measure semantic change (Figure \ref{measures}). 

\subsubsection*{Global measure}

The first measure analyzes global shifts in a word's vector semantics and is identical to the measure used in most previous works \cite{gulordava_distributional_2011,jatowt_framework_2014,kim_temporal_2014,hamilton_diachronic_2016}.
We simply take a word's vectors for two consecutive decades and measure the cosine distance between them, i.e.
\begin{equation}
	d^{G}(w^{(t)}_i, w^{(t+1)}_i) = \textrm{cos-dist}(\mb{w}^{(t)}_i,\mb{w}^{(t+1)}_i).
\end{equation}

\subsubsection*{Local neighborhood measure}

The second measure is based on the intuition that only a word's nearest semantic neighbors are relevant. 
For this measure, we first find word $w_i$'s set of $k$ nearest-neighbors (according to cosine-similarity) within each decade, which we denote by the ordered set $\mathcal{N}_k(w^{(t)}_i)$.
Next, to measure the change between decades $t$ and $t+1$, we compute a ``second-order'' similarity vector for $w^{(t)}_i$ from these neighbor sets with entries defined as
\begin{multline}
\mb{s}^{(t)}(j) = \textrm{cos-sim}(\mb{w}^{(t)}_i, \mb{w}^{(t)}_j)\\ \:\: \forall w_j \in \mathcal{N}_k(w^{(t)}_i) \cup \mathcal{N}_k(w^{(t+1)}_i),
\end{multline}
and we compute an analogous vector for $w_i^{(t+1)}$.
The second-order vector, $\mb{s}^{(t)}_i$, contains the cosine similarity of $\mb{w}_i$ and the vectors of all $w_i$'s nearest semantic neighbors in the the time-periods $t$ and $t+1$. 
Working with variants of these second-order vectors has been a popular approach in many  recent works, though most of these works define these vectors against the full vocabulary and not just a word's nearest neighbors \cite{del_prado_martin_case_2016,eger_linearity_2016,rodda_panta_2016}.

Finally, we compute the local neighborhood change as 
\begin{equation}\label{loc}
	d^{L}(w^{(t)}_i, w^{(t+1)}_i) = \textrm{cos-dist}(\mb{s}^{(t)}_i,\mb{s}^{(t+1)}_i).
\end{equation}	
This measures the extent to which $w_i$'s similarity with its nearest neighbors has changed. 

The local neighborhood measure defined in \eqref{loc} captures strong shifts in a word's paradigmatic relations but is less sensitive to global shifts in syntagmatic contexts \cite{schutze_vector_1993}.   
We used $k=25$ in all experiments (though we found the results to be consistent for $k\in[10,50]$). 

\subsection{Statistical methodology}

\begin{table}
\centering
\begin{tabular}{lcc} 
\toprule  
Dataset & \# Nouns &  \# Verbs \\
\midrule 
Google English All & 5299 & 2722\\
Google English Fic. & 4941 & 3128\\
German & 5443 &1844\\
French & 2310 & 4992\\
COHA (Word) & 4077 & 1267\\
COHA (Lemma) &  3389 & 783\\
\bottomrule
\end{tabular} 
\vspace{5pt}
\caption{\textbf{Number of nouns and verbs tested in each dataset.}}
\vspace{0pt}
\end{table}

To test whether nouns or verbs change more according to our two measures of change, we build on our previous work and used a linear mixed model approach \cite{hamilton_diachronic_2016}. 
This approach amounts to a linear regression where the model also includes ``random'' effects to account for the fact that the measurements for individual words will be correlated across time \cite{mcculloch_generalized_2001}. 

We ran two regressions per datatset: one with the global $d^{G}$ values as the dependent variables (DVs) and one with the local neighborhood $d^{L}$ values.
In both cases we examined the change between all consecutive decades and normalized the DVs to zero-mean and unit variance. 
We examined nouns/verbs within the top-10000 words by frequency rank and removed all words that occurred ${<}500$ times in the smaller COHA dataset. 
The independent variables are word frequency, the decade of the change (represented categorically), and variable indicating whether a word is a noun or a verb (proper nouns are excluded, as in \citealt{hamilton_diachronic_2016}).\footnote{Frequency was included since it is known to strongly influence the distributional measures \cite{hamilton_diachronic_2016}.} 

\begin{table*}
{\small
\centering
\begin{tabular}{lll}
\toprule
Word & 1850s context & 1990s context\\
\midrule
actually & ``...dinners which you have \underline{actually} eaten.''  & ``With that, I \underline{actually} agree.''\\
must & ``O, George, we \underline{must} have faith.'' & ``Which you \underline{must} have heard ten years ago...''\\
promise & ``I \underline{promise} to pay you...' & ``...the day \underline{promised} to be lovely.''\\
\midrule
gay & ``\underline{Gay} bridals and other merry-makings of men.'' & ``...the result of \underline{gay} rights demonstrations.''\\
virus & ``This young man is...infected with the \underline{virus}.'' & ``...a rapidly spreading computer \underline{virus}.''\\
cell &``The door of a gloomy \underline{cell}...'' & ``They really need their \underline{cell} phones.''\\
\bottomrule
\end{tabular}
}
\caption{\textbf{Example case-studies of semantic change.} The first three words are examples of regular linguistic shifts, while the latter three are examples of words that shifted due to exogenous cultural factors. Contexts are from the COHA data \protect\cite{davies_corpus_2010}.}
\label{case-words}
\vspace{-10pt}
\end{table*}
\begin{figure*}
\centering
\includegraphics[scale=0.9]{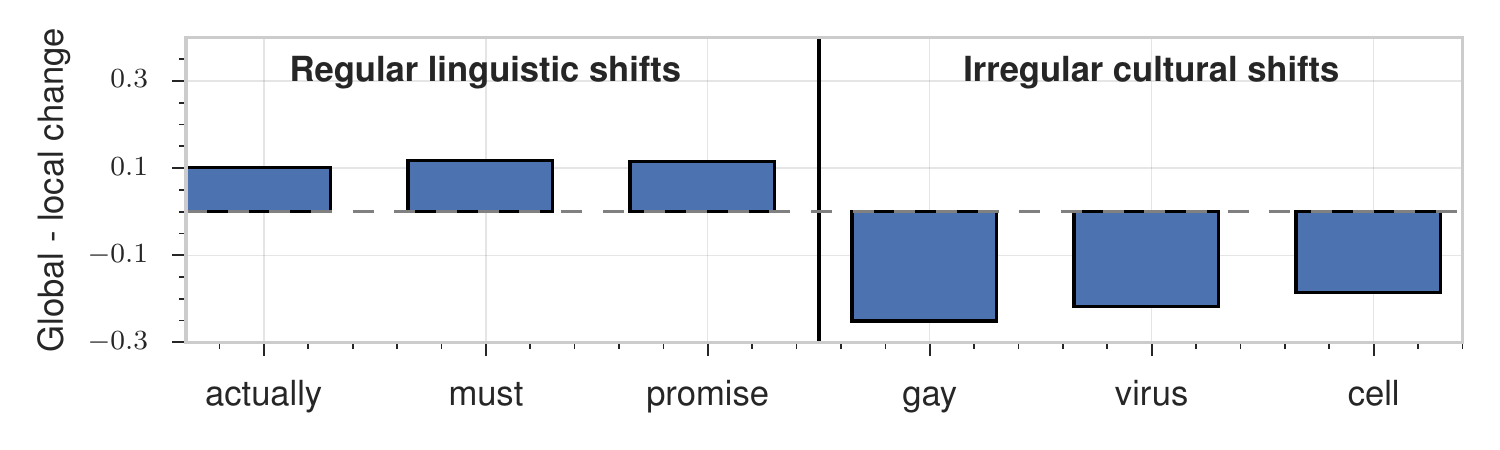}
\vspace{-5pt}
\caption{\textbf{The global measure captures classic examples of linguistic drift while the local measure captures example cultural shifts.} Examining the semantic distance between the 1850s and 1990s shows that the global measure is more sensitive to regular shifts (and vice-versa for the local measure). The plot shows the difference between the measurements made by the two methods.}
\label{case-studies}
\vspace{-10pt}
\end{figure*}

\section{Results}

Our results show that the two seemingly related measures actually result in drastically different notions of semantic change.  

\subsection{Nouns vs.\@ verbs}

The local neighborhood measure assigns far higher rates of semantic change to nouns across all languages and datasets while the opposite is true for the global distance measure, which tends to assign higher rates of change to verbs (Figure \ref{pos-effects}). 

We focused on verbs vs.\@ nouns since they are the two major parts-of-speech and previous research has shown that verbs are more semantically mutable than nouns and thus more likely to undergo linguistic drift \cite{gentner_verb_1988}, while nouns are far more likely to change due to cultural shifts like new technologies \cite{traugott_regularity_2001}.
However, some well-known regular linguistic shifts include rarer parts of speech like adverbs (included in our case studies below). 
Thus we also confirmed that the differences shown in Figure \ref{pos-effects} also hold when adverbs and adjectives are included along with the verbs.
This modified analysis showed analogous significant trends, which fits with previous research arguing that adverbial and adjectival modifiers are also often the target of regular linguistic changes \cite{traugott_regularity_2001}. 

The results of this large-scale regression analysis show that the local measure is more sensitive to changes in the nominal domain, a domain in which change is known to be driven by cultural factors. 
In contrast, the global measure is more sensitive to changes in verbs, along with adjectives and adverbs, which are known to be the targets of many regular processes of linguistic change \cite{traugott_regularity_2001,hopper_grammaticalization_2003}

\subsection{Case studies}

We examined six case-study words grouped into two sets. 
These case studies show that three examples of well-attested regular linguistic shifts (set A) changed more according to the global measure, while three well-known examples of cultural changes (set B) change more according to the local neighborhood measure. 
Table \ref{case-words} lists these words with some representative historical contexts \cite{davies_corpus_2010}. 

Set A contains three words that underwent attested regular linguistic shifts detailed in \newcite{traugott_regularity_2001}: \textit{actually}, \textit{must}, and \textit{promise}.
These three words represent three different types of regular linguistic shifts:
\textit{actually} is a case of subjectification (detailed in Figure \ref{measures}); \textit{must} shifted from a deontic/obligation usage (``you must do X'') to a epistemic one (``X must be the case''), exemplifying a regular pattern of change common to many modal verbs; and \textit{promise} represents the class of shifting ``performative speech acts'' that undergo rich changes due to their pragmatic uses and subjectification \cite{traugott_regularity_2001}. 
The contexts listed in Table \ref{case-words} exemplify these shifts.  

Set B contains three words that were selected because they underwent well-known cultural shifts over the last 150 years: \textit{gay}, \textit{virus}, and \textit{cell}.
These words gained new meanings due to uses in community-specific vernacular (\textit{gay}) or technological advances (\textit{virus}, \textit{cell}). 
The cultural shifts underlying these changes in usage --- e.g., the development of the mobile ``cell phone'' --- were unpredictable in the sense that they were not the result of regularities in human linguistic systems. 

Figure \ref{case-studies} shows how much the meaning of these word changed from the 1850s to the 1990s according to the two different measures on the English Google data.
We see that the words in set A changed more when measurements were made using the global measure, while the opposite holds for set B.

\section{Discussion}

Our results show that our novel local neighborhood measure of semantic change is more sensitive to changes in nouns, while the global measure is more sensitive to changes in verbs.
This mapping aligns with the traditional distinction between irregular cultural shifts in nominals and more regular cases of linguistic drift \cite{traugott_regularity_2001} and is further reinforced by our six case studies.

This finding emphasizes that researchers must develop and use measures of semantic change that are tuned to specific tasks.
For example, a cultural change-point detection framework would be more successful using our local neighborhood measure, while an empirical study of grammaticalization would be better off using the traditional global distance approach. 
Comparing measurements made by these two approaches also allows researchers to assess the extent to which semantic changes are linguistic or cultural in nature.

\section*{Acknowledgements} 
The authors thank C.\@ Manning, V.\@ Prabhakaran, S.\@ Kumar, and our anonymous reviewers for their helpful comments.
This research has been supported in part by NSF
CNS-1010921,     
IIS-1149837, IIS-1514268
NIH BD2K,
ARO MURI, DARPA XDATA,
DARPA SIMPLEX,
Stanford Data Science Initiative,
SAP Stanford Graduate Fellowship, NSERC PGS-D,
Boeing,          
Lightspeed,			
and Volkswagen.  
\bibliography{cultling_change}
\bibliographystyle{emnlp2016}

\end{document}